\documentclass[sigconf]{acmart}
\usepackage{graphicx}

\acmConference[CODS-COMAD '25]{8th ACM IKDD CODS and 30th COMAD}{December 17--20, 2025}{Pune, India}
\acmBooktitle{8th ACM IKDD CODS and 30th COMAD (CODS-COMAD '25), December 17--20, 2025, Pune, India}

\usepackage{multirow}
\usepackage{hhline}
\usepackage{booktabs}
\usepackage{color}
\usepackage{array}
\usepackage{ragged2e}
\usepackage{tabularray}
\usepackage{longtable}
\usepackage{color}
\usepackage{tabularray}
\usepackage{float}
\usepackage[normalem]{ulem}
\usepackage{colortbl}
\usepackage{hhline}
\usepackage{amsmath} 
\usepackage{tcolorbox}
\usepackage{url}

\title{Adapting Multilingual Models to Code-Mixed Tasks via Model Merging}

\author{Prashant Kodali}
\affiliation{
  \institution{IIIT Hyderabad}
  \city{Hyderabad}
  \country{India}
}
\email{prashant.kodali@research.iiit.ac.in}

\author{Vaishnavi Shivkumar}
\affiliation{
  \institution{IIIT Hyderabad}
  \city{Hyderabad}
  \country{India}
}
\email{vaishnavi.shivkumar@students.iiit.ac.in}

\author{Swarang Joshi}
\affiliation{
  \institution{IIIT Hyderabad}
  \city{Hyderabad}
  \country{India}
}
\email{swarang.joshi@research.iiit.ac.in}
\author{Monojit Choudhury}
\affiliation{
  \institution{MBZUAI}
  \city{Abu Dhabi}
  \country{UAE}
}
\email{monojit.choudhury@mbzuai.ac.ae}

\author{Ponnurangam Kumaraguru}
\affiliation{
  \institution{IIIT Hyderabad}
  \city{Hyderabad}
  \country{India}
}
\email{pk.guru@iiit.ac.in}

\author{Manish Shrivastava}
\affiliation{
  \institution{IIIT Hyderabad}
  \city{Hyderabad}
  \country{India}
}
\email{m.shrivastava@iiit.ac.in}

\begin{document}

\begin{abstract}

We study model merging as a practical alternative to conventional adaptation strategies for code-mixed NLP. Starting from a multilingual base model, we: (i) perform continued pre-training (CPT) on unlabeled code-mixed text to obtain an adapted checkpoint, (ii) merge checkpoint with the base model, and (iii) fine-tune~(FT) on the downstream task data. We evaluate our approach for sentence classification (sentiment and hate speech) task in English-Hindi (En-Hi) and English-Spanish (En-Es) using XLM-R and Llama-3.2-1B models. Our results show that merged models consistently outperform full fine-tuning and CPT$\to$FT. We observe gains of 2--5 points in $F_1$ over full fine-tuning and $\sim$1--2 points over CPT$\to$FT, indicating that unlabeled data is leveraged more effectively via merging than via CPT alone. Zero-/few-shot prompting with larger LLMs (e.g., Llama-3.3-70B) lags behind fine-tuned and merged checkpoints, underscoring limits of in-context learning for code-mixed inputs. We further test cross-pair transfer by training on En--Hi and evaluating on En--Ta and En--Ml: merged checkpoints transfer more strongly than monolingual-English baselines (e.g., TV/TIES variants reaching 0.65--0.68 $F_1$ vs.\ 0.61--0.63 for full fine-tuning), suggesting that code-mixed knowledge is a more reliable substrate for low-resource pairs. We conclude with adaptation recipes matched to common data regimes (labeled only; labeled+unlabeled; transfer-only) and discuss limitations and scaling considerations for broader tasks and larger models.
\end{abstract}

\ccsdesc[500]{Computing methodologies~Natural language processing}


\keywords{LLMs, Multilingual NLP, Code-Mixing}

\maketitle
\settopmatter{authorsperrow=1} 
\renewcommand{\shortauthors}{} 
\section{Introduction} \label{sec:01_intro}

Code-mixed text—where two or more languages appear within a single utterance or speech event~\cite{gumperz1977sociolinguistic}—poses persistent challenges for natural language processing (NLP) systems~\cite{dogruoz-etal-2021-survey, zhang2023multilingual, winata-etal-2023-decades}. A common response has been to fine-tune large multilingual pre-trained models on task-specific datasets for code-mixed inputs~\cite{khanuja2020gluecos, aguilar-etal-2020-lince}. While effective, this approach depends on the availability of high-quality annotations and on general-purpose multilingual models having seen sufficient code-mixed data during pre-training to capture the characteristics of code-mixed language.

A complementary strategy for adapting language models to new domains or linguistic settings is continued pre-training (CPT) on unlabeled data~\cite{gururangan-etal-2020-dont, liu-etal-2021-continual, ruder2021lmfine-tuning}. Model merging~\cite{yang2024modelmergingllmsmllms} has emerged as a promising approach for adapting models to new domains, tasks, and languages, while mitigating several limitations of CPT. Although CPT can aid domain adaptation, it can also erode general language understanding~\cite{10.1007/978-3-031-70563-2_6,alexandrov-etal-2024-mitigating}. In contrast, recent work shows that model merging can improve task performance~\cite{pmlr-v162-wortsman22a,choshen2022fusingfinetunedmodelsbetter,Izmailov2018AveragingWL}, enhance robustness and out-of-domain generalization~\cite{pmlr-v162-wortsman22a,jin2023dataless}, and enable multitask models~\cite{yadav2023tiesmerging, ilharco2023editing}.

We argue that model merging is well-suited to code-mixed settings because it can incorporate code-mixed knowledge while preserving strong monolingual representations. Since code-mixed text contains monolingual spans from its constituent languages (L1, L2) and coexists with purely monolingual utterances, maintaining these capabilities is crucial for downstream performance. By comparison, CPT$\rightarrow$FT risks degrading monolingual processing during adaptation. Our hypothesis is that merging a base multilingual model with a checkpoint adapted on a modest code-mixed corpus can outperform traditional strategies by combining strong monolingual representations with newly acquired code-mixing capabilities, yielding a more robust solution for code-mixed text.

Model merging integrates task- or domain-specific models into a unified composite, aggregating knowledge from all merged sources. In resource-constrained code-mixed settings, leveraging monolingual or other code-mixed resources can further improve performance. While model merging offers a modular mechanism for combining diverse data sources, its effectiveness for code-mixed tasks remains underexplored and merits closer study. 

Beyond improving performance for a single code-mixed language pair, a natural question is whether gains from model merging transfer to other language pairs performing the same downstream task. In such cases, the task formulation remains fixed, but the language composition changes, introducing new patterns of mixing, grammar, and vocabulary overlap. These shifts create both challenges and opportunities: the model must adjust to different lexical and syntactic cues, yet shared structures or typological similarities may support better generalization. Understanding how language composition affects transfer is especially important in low-resource settings, where direct annotations for the target pair may be unavailable and cross-pair transfer becomes a viable adaptation strategy.

Given this context, we pose two primary research questions: \textbf{RQ 1:} Can model merging serve as an alternative to fine-tuning approaches for code-mixed tasks? \textbf{RQ 2:} Are model merging methods effective for integrating capabilities from multiple data sources that differ in language composition and training supervision?

This work examines model merging for code-mixed sentence classification, an approach not previously explored in code-mixed settings. We run experiments on English–Hindi and English–Spanish datasets using XLM-R, and Llama 3.2 1B to assess how major multilingual models respond to model merging. Concretely, we merge a continued pre-trained checkpoint with a base model and then fine-tune for code-mixed tasks. We also evaluate model merging under different data availability scenarios to test its ability to incorporate both monolingual and code-mixed resources.

Our main findings are as follows:
\begin{enumerate}
    \item We introduce model merging as an adaptation strategy for code-mixed NLP tasks, distinct from conventional fine-tuning and continued pre-training, and observe improved downstream performance across language pairs, models, and tasks.
    \item For cross-lingual transfer, merged models trained on code-mixed resources outperform those trained on monolingual English data, with the advantage most pronounced for low-resource language pairs, indicating that code-mixed knowledge is a stronger basis for adaptation when target-pair data is scarce.
\end{enumerate}

\section{Computational Approaches to Code-Mixed Text}\label{sec:02_background}

Modern work on code-mixed NLP primarily adapts \textit{multilingual} pretrained models (e.g., mBERT, XLM-R)~\cite{pires-etal-2019-multilingual, muller2020multilinguallanguagemodelstransfer, tan-joty-2021-code, srivastava-singh-2022-hinglisheval, winata-etal-2023-decades, Winata2021AreMM, rani-etal-2024-findings, 10.1145/3503162.3503177}. Such models are adapted via task-specific fine-tuning and their efficacy has been shown on benchmarks such as LinCE and GLUECoS, which span token-level and sentence-level tasks across pairs like En–Hi and En–Es~\cite{aguilar-etal-2020-lince, khanuja2020gluecos}. In parallel, continued pre-training (CPT) on unlabeled code-mixed corpora has emerged as a complementary strategy before downstream fine-tuning~\cite{das-etal-2023-improving}. While both routes are effective, they rely on labeled and/or unlabeled code-mixed resources that remain scarce or unevenly distributed across language pairs, and they typically optimize a \emph{single} model instance rather than composing multiple models.

A less-explored axis concerns leveraging \textit{monolingual} resources for code-mixed tasks. Prior work has examined translating mixed inputs to constituent languages~\cite{pant-dadu-2020-towards}, using language-specific models, or building meta-embeddings~\cite{Winata2021AreMM} and non-Transformer encoders~\cite{aguilar-solorio-2020-english}; these can yield gains in select settings but do not directly address the need to \emph{compose} capabilities across sources. In contrast, \emph{modular} adaptation by \emph{model merging}---combining parameters from models specialized for different data or tasks (e.g., Task Arithmetic and TIES variants)---offers a data- and compute-efficient alternative that can, in principle, integrate code-mixed competence with strong monolingual representations without access to original training data.

\paragraph{Modular Methods for Model Adaptation}
Model merging is a powerful technique that combines the parameters of different models, creating a composite model with enhanced functionality, all without the need for original training data or heavy computational resources. Methods like Task Arithmetic~\cite{ilharco2023editing} and its variants~\cite{yadav2023tiesmerging, yu2024languagemodelssupermario} demonstrate the potential to merge models with diverse capacities to achieve a unified function. This concept can be extended to code-mixed tasks by combining task- or domain-specific modules. Exploring the use of model merging methods using monolingual or unlabeled resources, presents a promising avenue for advancing code-mixed NLP.

To our knowledge, \textbf{model merging has not been applied to code-mixed NLP} prior to this line of work. Framing code-mixed adaptation as a modular composition problem allows us to ask: when labeled code-mixed data is limited, can we merge a base multilingual model with a CPT-adapted checkpoint (or with monolingual task vectors) and then fine-tune for the target task? This perspective also clarifies cross-lingual transfer: merged checkpoints trained with code-mixed resources tend to transfer more reliably to new code-mixed pairs than those relying solely on monolingual English supervision, highlighting that code-mixed signals are especially valuable under low-resource target conditions.

\section{Methodology}\label{methodology}\label{sec:03_methodology}
 

\textbf{Notations} Let $T$ represent the target task. Let $D^T_{L1-L2}$ stand for a code-mixed dataset for task $T$ in the L1-L2 language pair (e.g., En-Hi), and ~($D^T_{L1}$) for a monolingual dataset in L1 for the same task $T$. A language model can be adapted to code-mixed data by performing continued pre-training using an unlabeled code-mixed text corpus ~($D^{CM}_{L1-L2}$). We start with a base pre-trained model ~($\theta$), and when fine-tuned on the downstream task $T$, we get $\theta^T$. If $\theta$ undergoes continued pre-training on a code-mixed dataset, it results in the ~$\theta^{CM}_{L1-L2}$ model, which is expected to better handle code-mixed text.

\textbf{Baselines}: To analyze various trade-offs and ensure a fair evaluation, we use full-model fine-tuning (FullFT) as our primary baseline.

\subsection{Continued Pre-Training}
Adapting a language model to a new domain or task has been shown to improve performance~\cite{gururangan-etal-2020-dont}. We adapt a language model to code-mixed corpus, and such treatment can lead to improved downstream task performance. We use existing unlabeled code-mixed corpus for continued pre-training of base model using Masked Language Modeling objective for Encoder-only models, and Causal LM objective for Decoder-only models~(refer details Section \ref{sec:04_exp_setup} for more details). 
We take the continued pre-training checkpoints and perform fine-tuning on the downstream task dataset. We refer this method as ``CPT$\rightarrow$FT''.  We should note that domain adaptation incurs additional computational cost, which is much higher than the fine-tuning.

\subsection{Model Merging}\label{sec:04_methodology_model_merging}
For model merging we consider two methods - a) Task Arithmetic ~\cite{ilharco2023editing}; b) TIES (TrIm, Elect, Sign) ~\cite{yadav2023tiesmerging}. Task Arithmetic defines a direction, or Task Vector, in a model’s weight space that enhances task performance when followed. These vectors, calculated by subtracting pre-trained model weights from fine-tuned ones, guide neural network behavior. Task vectors can be modified and combined using operations like addition or negation to adjust model behavior. TIES is a variation that addresses interference from redundant parameters and disagreements in parameter signs across models.

We compute a task vector for continued pre-training of $\theta$ on code-mixed text ($\tau_{CPT}$) (Eq~~\ref{eq:task_vector_cpt}). We add one Task Vector at a time and fine-tune task specific dataset ($D^T_{enhi}$).
\begin{gather}
\theta^T_{CPT \rightarrow L1-L2} = f(\theta  \oplus \lambda \tau_{CPT})\label{eq:task_vector_cpt}
\end{gather}
where $f$ denotes the fine-tuning of the model on a downstream task, and $\lambda$, scaling term is determined using held-out validation sets.


 In TIES, the resulting model weights are not a simple weighted average of all parameters; instead, they are adjusted based on the magnitude and weight of the task vectors. For further discussion, 
 we refer to Eq.~\ref{eq:task_vector_cpt} as ``TV''.
 In case of TIES they are referred to as ``TIES''.

\subsection{Leveraging labeled monolingual data}\label{sec:04_methodology_en_encpt}
We leverage monolingual English data in our adaptation methods and evaluate on the code-mixed task.
To compare cross-lingual/cross-dataset transfer we consider two baselines- a) Joint Training~(JointFT): English and English-Hindi task datasets are combined and model is fine-tuned on the combined dataset; b) Sequential Training~(SeqFT): we first train the model on English task dataset, followed by fine-tuning on the English-Hindi code-mixed dataset. Additionally, we look at the following:


\paragraph{Model Merging.}
Similarly, we compute a task vector $\tau^T_{en}$ for the English task:
\begin{gather}
\theta^T_{en \rightarrow enhi} = f(\theta  \oplus \lambda \tau^T_{en})\label{eq:task_vector_en}
\end{gather}
where $f$ denotes fine-tuning on $D^T_{enhi}$ and $\lambda$ is a scaling factor. We also merge $\tau^T_{en}$ and $\tau_{CPT}$ before fine-tuning:
\begin{equation} \label{eq:task_vector_combined}
    \theta^T_{\left\{ en + CPT\right\} \rightarrow enhi} = f(\theta  \oplus   \lambda_1\tau^T_{en} \oplus  \lambda_2 \tau_{CPT})
\end{equation}
When using TIES, the resulting model weights are adjusted based on parameter magnitudes and signs rather than simple averaging.

We will be refering to these models as~``TV-$T_{en}\rightarrow$FT - only en (Eq.~\ref{eq:task_vector_en}), etc.

\begin{table*}[!h]
\centering
\caption{Statistics of all the datasets used in our study.}

\resizebox{0.85\linewidth}{!}{%
\begin{tabular}{cllccc} 
\hline\hline
\textbf{Language Pair}               & \multicolumn{1}{c}{\textbf{Task}}                                                                & \multicolumn{1}{c}{\textbf{Dataset}} & \textbf{Train}                                                                   & \textbf{Dev}                                                                  & \textbf{Test}                                                                   \\ 
\hline\hline
\multirow{5}{*}{English-Hindi}       & \multicolumn{1}{c}{\multirow{3}{*}{\begin{tabular}[c]{@{}c@{}}Sentiment\\Analysis\end{tabular}}} & \citet{khanuja2020gluecos}                              & \begin{tabular}[c]{@{}c@{}}7937\\Neu: 3583\\Pos : 2543\\Neg: 1811\end{tabular}   & \begin{tabular}[c]{@{}c@{}}1134\\Neu: 512\\Pos : 349\\Neg: 273\end{tabular}   & \begin{tabular}[c]{@{}c@{}}2268\\Neu: 1041\\Pos : 709\\Neg: 518\end{tabular}    \\ 
\cline{3-6}
                                     & \multicolumn{1}{c}{}                                                                             & \citet{patwa-etal-2020-semeval}                     & \begin{tabular}[c]{@{}c@{}}14000\\Neu:~ 5264\\Pos : 4634\\Neg: 4102\end{tabular} & \begin{tabular}[c]{@{}c@{}}3000\\Neu:~ 1128\\Pos : 982\\Neg: 890\end{tabular} & \begin{tabular}[c]{@{}c@{}}3000\\Neu:~ 1100\\Pos : 1000\\Neg: 900\end{tabular}  \\ 
\cline{3-6}
                                     & \multicolumn{1}{c}{}                                                                             & \citet{Prabhu2016TowardsSL}                  & \begin{tabular}[c]{@{}c@{}}2715\\Neu:~ 411\\Pos : 936\\Neg: 1368\end{tabular}    & \begin{tabular}[c]{@{}c@{}}388\\Neu:~ 51\\Pos : 152\\Neg: 185\end{tabular}    & \begin{tabular}[c]{@{}c@{}}776\\Neu: 404\\Pos : 264\\Neg: 404\end{tabular}      \\ 
\cline{2-6}
                                     & Hate Speech                                                                                      & \citet{bohra-etal-2018-dataset}              & \begin{tabular}[c]{@{}c@{}}3202\\Hate:~ 1163\\Non-Hate: 2039\end{tabular}        & \begin{tabular}[c]{@{}c@{}}458\\Hate:~ 179\\Non-Hate: 279\end{tabular}        & \begin{tabular}[c]{@{}c@{}}915\\Hate:~ 319\\Non-Hate: 956\end{tabular}          \\ 
\cline{2-6}
                                     & Unlabeled Corpus                                                                                 & \citet{das-etal-2023-improving}              & \multicolumn{1}{l}{166464}                                                       & \multicolumn{1}{l}{~-~}                                                       & \multicolumn{1}{l}{18495}                                                       \\ 
\hline
\multicolumn{1}{l}{}                 &                                                                                                  &                                      &                                                                                  &                                                                               &                                                                                 \\ 
\hline
\multirow{3}{*}{English-Spanish}     & \multicolumn{1}{c}{\multirow{2}{*}{\begin{tabular}[c]{@{}c@{}}Sentiment\\Analysis\end{tabular}}} & \citet{khanuja2020gluecos}                              & \begin{tabular}[c]{@{}c@{}}1195\\Neu: 571\\Pos : 337\\Neg: 287\end{tabular}      & \begin{tabular}[c]{@{}c@{}}171\\Neu: 66\\Pos : 48\\Neg: 57\end{tabular}       & \begin{tabular}[c]{@{}c@{}}342\\Neu: 157\\Pos : 85\\Neg: 90\end{tabular}        \\ 
\cline{3-6}
                                     & \multicolumn{1}{c}{}                                                                             & \citet{patwa-etal-2020-semeval}                     & \begin{tabular}[c]{@{}c@{}}9836\\Neu: 2727\\Pos : 5511\\Neg: 1598\end{tabular}   & \begin{tabular}[c]{@{}c@{}}1406\\Neu:~ 369\\Pos : 784\\Neg: 253\end{tabular}  & \begin{tabular}[c]{@{}c@{}}2811\\Neu: 784\\Pos : 1590\\Neg: 437\end{tabular}    \\ 
\cline{2-6}
                                     & Unlabeled Corpus                                                                                 & \citet{das-etal-2023-improving}              & 52940                                                                            & -                                                                             & 6617                                                                            \\ 
\hline
\multicolumn{1}{l}{English-Tamil}    & Sentiment Analysis                                                                               & \citet{chakravarthi-etal-2020-corpus}        & -~                                                                               & -~                                                                            & \begin{tabular}[c]{@{}c@{}}1231\\Neu: 209\\Pos: 287\\Neg: 165\end{tabular}      \\ 
\hline
\multicolumn{1}{l}{English-Malyalam} & Sentiment Analysis                                                                               & \citet{chakravarthi-etal-2020-sentiment}     & -                                                                                & -                                                                             & \begin{tabular}[c]{@{}c@{}}480\\Neu: 205\\Pos: 224\\Neg: 51\end{tabular}        \\ 
\hline
\multicolumn{1}{l}{English}          & Sentiment Analysis                                                                               & SST - 5~\cite{socher-etal-2013-recursive}                             & \begin{tabular}[c]{@{}c@{}}8544\\Neu: 1624\\Pos: 3610\\Neg: 3310\end{tabular}    & \begin{tabular}[c]{@{}c@{}}1101\\Neu: 229\\Pos: 444\\Neg: 428\end{tabular}    & \begin{tabular}[c]{@{}c@{}}2210\\Neu: 389\\Pos: 909\\Neg: 912\end{tabular}      \\
\hline\hline
\end{tabular}
}
\label{tab:dataset_stats}

\end{table*}

\section{Experimental Setup}\label{sec:04_exp_setup}
\textbf{Models:} We use multilingual models, as their fine-tuning has consistently shown strong results \cite{khanuja2020gluecos, das-etal-2023-improving}.
For our experiments, we use mBERT~\cite{devlin-etal-2019-bert} and XLM-R~\cite{conneau-etal-2020-unsupervised}, which are widely used in code-mixed studies, along with Llama 3.2 1B model\footnote{\url{https://huggingface.co/meta-llama/Llama-3.2-1B}}.
Due to computational constraints, our model adaptation methods are restricted to Llama 1B; however, we conduct inference on larger models including Llama 3.2 3B, Llama 3.1 8B, and Llama 3.3 70B.

\noindent\textbf{Datasets:} Our study focuses on English-Hindi (En-Hi) and English-Spanish (En-Es) code-mixed sentiment classification tasks, as they are part of code-mixing benchmarks GLUECoS~\cite{khanuja2020gluecos} and LinCE~\cite{aguilar-etal-2020-lince}. For En-Hi we consider two tasks - a) Sentiment Analysis using three datasets — GLUECoS, Sentimix~\cite{patwa-etal-2020-semeval}, and~\citet{Prabhu2016TowardsSL}—all featuring 3-class labels (positive, neutral, negative); b) Hate Speech Classification~\cite{bohra-etal-2018-dataset}. For En-Es, we carry out experiments on Sentiment Analysis task, and use two datasets released by \citet{patwa-etal-2020-semeval} and GLUECoS benchmark. For transfer from monolingual English task dataset, we use English SST5 dataset~\cite{socher-etal-2013-recursive}.

 To evaluate cross lingual to new language pairs, we consider three target datasets - a) English-Tamil~(En-Ta)~\cite{chakravarthi-etal-2020-corpus}; b) English-Malayalam~(En-Ml)~\cite{chakravarthi-etal-2020-sentiment}; c) English-Spanish~(En-Es) from GLUECoS benchmark. For unlabeled code-mixed corpus, we use the dataset released by ~\citet{das-etal-2023-improving} for both English-Hindi and English-Spanish. 

\noindent\textbf{Zero-shot and Few-shot Prompting on Llama:}
We prompt Llama-3.2-1B, Llama-3.2-3B, Llama-3.1-8B, and Llama-3.3-70B to perform the given tasks. We evaluate them using 0, 5, 10, 15 and 20-shot setting.  





\textbf{Datasets:} Table~\ref{tab:dataset_stats} provides statistics of all the datasets we have used in our experiments, along with the label wise distributions in all splits. We include all the labeled, unlabeled datasets in the table for ready reference. SST5 dataset~\cite{socher-etal-2013-recursive} has 5 classes, and to use in our experiments we convert its 5-class labels into 3 classes by merging ``very positive'' with ``positive'' and ``very negative'' with ``negative''. Wherever all three splits (train-test-valid) are available we use the same, and in case they are not available we create 70-20-10 split from the available samples. We use F1 as the measure of performance across all the  data and training configurations. F1 is suitable as there is some dataset imbalances across code-mixed datasets.

\textbf{Prompts:} We used the following prompts for our zero and few shot experiments. For few-shot prompt, we include the examples and the corresponding labels in between the instruction and the input sample. For the k-shot settings, we carry out five runs, each run with different set of randomly sampled k examples from the train dataset, while ensuring that all the labels are well represented in the examples.

The zero shot prompt used for sentiment analysis is as follows:

\begin{tcolorbox}[colback=lightgray,coltext=black]
\begin{center}
    \textbf{Sentiment Analysis}
\end{center}
\begin{quote}
\texttt{You are a sentiment classifier. The sentiment of the text is either positive, negative, or neutral. Give a one-word response.}

\texttt{Text: [INPUT]}

\texttt{The sentiment of this text is:}
\end{quote}
\begin{center}
    \textbf{Hate Speech Classification}    
\end{center}
\begin{quote}
\texttt{"Do you think this comment is hate speech? Answer Yes or No.  \
        Give a 1-word answer. Comment: [INPUT]"}
\end{quote}

\end{tcolorbox}

\textbf{Hardware \& Hyperparameters:} We conduct all experiments on a combination of Nvidia 1080 Ti 12 GB GPUs, Nvidia A6000 32 GB GPU, using a single GPU. We use Nvidia A6000 for Llama experiemnts, and 1080Ti for all other experiments. Batch sizes range from 8 to 128, depending on the dataset and approach. Learning rates were selected after a search between [1e-2, 5e-5], with 1e-5 working best for full model training, 1e-4 for LoRA layers. All experiments ran for with early stopping~(stop if model performance doesn't change by 0.5 F1 point across 3 evaluations), maximum up to 20 epochs. We use AdamW optimizer with default optimizer parameters for all our experiments. We implemented the methods using PyTorch, Huggingface Transformers~\cite{wolf-etal-2020-transformers}, PEFT~\cite{peft}, and mergekit~\cite{goddard2024arcee} for model merging methods. Upon publication code, data, models, and detailed hyperparameters configurations will be publicly released for reproducibility.

\begin{table*}
\centering
\caption{Comparison of fine-tuning, continued pretraining, and model merging strategies for multilingual \textbf{sentiment analysis} (En--Hi, En--Es) and \textbf{hate speech detection} (En--Hi, En--Es) across multiple benchmarks (GLUEcos, SentiMix, Prabhu \textit{et al.}, Bohra \textit{et al.}). Results are reported for XLM-R and LLaMA~2~13B with full fine-tuning, sequential and joint training, transfer variants (TV, TIES), and combinations with continued pretraining (CPT). Zero- and few-shot prompting results with LLaMA~2~70B are also shown for comparison. \textbf{Main takeaway:} Parameter-efficient strategies such as TIES and transfer variants, particularly when combined with CPT, achieve competitive or superior performance to full fine-tuning, highlighting their effectiveness for code-mixed and multilingual tasks under resource constraints.}
\label{tab:main_results}
\resizebox{0.79\linewidth}{!}{%
\begin{tabular}{llcccccc} 
\hline\hline
\multicolumn{2}{l}{\multirow{3}{*}{\textbf{Methodology}}}                                                                                                                                                                                                                                                                                                                                                & \multicolumn{5}{c}{\textbf{Sentiment Analysis}}                                                                                                                                                                                                                                                                                                                                                                                                                         & \multicolumn{1}{l}{\textbf{Hatespeech}}                                                           \\ 
\cline{3-8}
\multicolumn{2}{l}{}                                                                                                                                                                                                                                                                                                                                                                                     & \multicolumn{3}{c}{\textbf{En-Hi}}                                                                                                                                                                                                                                & \multicolumn{2}{c}{\textbf{En-Es}}                                                                                                                                                                  & \textbf{En-Hi}                                                                                    \\ 
\cline{3-8}
\multicolumn{2}{l}{}                                                                                                                                                                                                                                                                                                                                                                                     & \multicolumn{1}{l}{\textbf{Gluecos}}                                           & \multicolumn{1}{l}{\textbf{Sentimix}}                                          & \multicolumn{1}{l}{\textbf{Prabhu et al}}                                                       & \multicolumn{1}{l}{\textbf{Gluecos}}                                                             & \multicolumn{1}{l}{\textbf{Sentimix}}                                                            & \multicolumn{1}{l}{\textbf{Bohra et al}}                                                          \\ 
\hline\hline
\multicolumn{8}{c}{\textbf{XML-R}}                                                                                                                                                                                                                                                                                                                                                                                                                                                                                                                                                                                                                                                                                                                                                                                                                                                                                                                                                     \\ 
\hline\hline
\multicolumn{2}{c}{\textbf{Full FT}}                                                                                                                                                                                                                                                                                                                                                                     & \begin{tabular}[c]{@{}c@{}}0.62\\(0.62 ± 0.008)\end{tabular}                   & \begin{tabular}[c]{@{}c@{}}0.64\\(0.63 ± 0.005)\end{tabular}                   & \begin{tabular}[c]{@{}c@{}}0.75\\(0.74 ± 0.015)\end{tabular}                                    & \begin{tabular}[c]{@{}c@{}}0.647\\(0.61 ± 0.035)\end{tabular}                                    & \begin{tabular}[c]{@{}c@{}}0.39\\(0.35 ± 0.021)\end{tabular}                                     & \begin{tabular}[c]{@{}c@{}}0.71\\(0.70 ± 0.001\end{tabular}                                       \\
\multicolumn{2}{c}{\textbf{CPT $\rightarrow$ FT}}                                                                                                                                                                                                                                                                                                                                                                      & \begin{tabular}[c]{@{}c@{}}0.65\\(0.64 ± 0.007)\end{tabular}                   & \begin{tabular}[c]{@{}c@{}}0.64\\(0.64 ± 0.005)\end{tabular}                   & \begin{tabular}[c]{@{}c@{}}0.76\\(0.76 ± 0.007)\end{tabular}                                    & \begin{tabular}[c]{@{}c@{}}0.66\\(0.64 ± 0.026)\end{tabular}                                     & \begin{tabular}[c]{@{}c@{}}0.57\\(0.54 ± 0.023)\end{tabular}                                     & \begin{tabular}[c]{@{}c@{}}0.70\\(0.69 ± 0.019)\end{tabular}                                      \\
\multicolumn{2}{c}{\textbf{Seq FT}}                                                                                                                                                                                                                                                                                                                                                                      & \begin{tabular}[c]{@{}c@{}}0.63\\(0.62 ± 0.006)\end{tabular}                   & \begin{tabular}[c]{@{}c@{}}0.63\\(0.63 ± 0.002)\end{tabular}                   & \begin{tabular}[c]{@{}c@{}}0.74\\(0.73±0.008)\end{tabular}                                      & \begin{tabular}[c]{@{}c@{}}0.65\\(0.64 ± 0.060)\end{tabular}                                     & \begin{tabular}[c]{@{}c@{}}0.40\\(0.39 ± 0.021)\end{tabular}                                     & \begin{tabular}[c]{@{}c@{}}0.70\\(0.70 ± 0.001\end{tabular}                                       \\
\multicolumn{2}{c}{\textbf{Joint FT}}                                                                                                                                                                                                                                                                                                                                                                    & \begin{tabular}[c]{@{}c@{}}0.63\\(0.62±0.009)\end{tabular}                     & \begin{tabular}[c]{@{}c@{}}0.64\\(0.63±0.010)\end{tabular}                     & \begin{tabular}[c]{@{}c@{}}0.72\\(0.71±0.010)\end{tabular}                                      & \begin{tabular}[c]{@{}c@{}}0.65\\(0.64 ± 0.060)\end{tabular}                                     & \begin{tabular}[c]{@{}c@{}}0.40\\(0.39 ± 0.01)\end{tabular}                                      & \begin{tabular}[c]{@{}c@{}}0.70\\(0.70 ± 0.003\end{tabular}                                       \\ 
\hline
\multirow{3}{*}{\textbf{TV }}                                                   & \textbf{\textbf{Base Model $\oplus  \tau_{CPT}$}}                                                                                                                                                                                                                                                                                     & \begin{tabular}[c]{@{}c@{}}0.65\\(0.64 ± 0.005)\end{tabular}                   & \begin{tabular}[c]{@{}c@{}}\textbf{0.66}\\\textbf{(0.64 ± 0.006)}\end{tabular} & \begin{tabular}[c]{@{}c@{}}\textbf{0.78}\\\textbf{(0.77 ± 0.006)}\end{tabular}                  & \begin{tabular}[c]{@{}c@{}}\textbf{0.67}\\\textbf{(0.64 ± 0.023)}\end{tabular}                   & \begin{tabular}[c]{@{}c@{}}0.56\\(0.54 ± 0.021)\end{tabular}                                     & \begin{tabular}[c]{@{}c@{}}\textbf{0.73}\\\textbf{(0.71 ± 0.016)}\end{tabular}                    \\
                                                                                & \textbf{\textbf{\textbf{\textbf{Baes Model $\oplus \tau_{en}$ }}}}                                                                                                                                                                                                                                                                   & \begin{tabular}[c]{@{}c@{}}0.63\\(0.63± 0.002)\end{tabular}                    & \begin{tabular}[c]{@{}c@{}}0.63\\(0.63± 0.008)\end{tabular}                    & \begin{tabular}[c]{@{}c@{}}0.75\\(0.74± 0.009)\end{tabular}                                     & \begin{tabular}[c]{@{}c@{}}0.65\\(0.65 ± 0.001)\end{tabular}                                     & \begin{tabular}[c]{@{}c@{}}0.56\\(0.55 ± 0.021)\end{tabular}                                     & \begin{tabular}[c]{@{}c@{}}0.71\\(0.71 ± 0.019)\end{tabular}                                      \\
                                                                                & \textbf{\textbf{\textbf{\textbf{\textbf{\textbf{\textbf{\textbf{Base Model $\oplus \tau_{CPT} \oplus \tau_{en}$ }}}}}}}}                                                                                                                                                                                                                         & \begin{tabular}[c]{@{}c@{}}0.65\\(0.64± 0.007)\end{tabular}                    & \begin{tabular}[c]{@{}c@{}}0.64\\(0.63± 0.009)\end{tabular}                    & \begin{tabular}[c]{@{}c@{}}\textbf{\textbf{0.78}}\\\textbf{\textbf{(0.77± 0.008)}}\end{tabular} & \begin{tabular}[c]{@{}c@{}}\textbf{\textbf{0.67}}\\\textbf{\textbf{(0.66 ± 0.023)}}\end{tabular} & \begin{tabular}[c]{@{}c@{}}\textbf{\textbf{0.58}}\\\textbf{\textbf{(0.57 ± 0.023)}}\end{tabular} & \begin{tabular}[c]{@{}c@{}}0.72\\(0.72 ± 0.003)\end{tabular}                                      \\ 
\hline
\multirow{3}{*}{\textbf{TIES}}                                                  & \textbf{\textbf{\textbf{\textbf{Base Model $\oplus \tau_{CPT}$}}}}                                                                                                                                                                                                                                                                   & \begin{tabular}[c]{@{}c@{}}0.64\\(0.63 ± 0.011)\end{tabular}                   & \begin{tabular}[c]{@{}c@{}}0.64\\(0.63 ± 0.004)\end{tabular}                   & \begin{tabular}[c]{@{}c@{}}0.77\\(0.76 ± 0.009)\end{tabular}                                    & \begin{tabular}[c]{@{}c@{}}0.66\\(0.64 ± 0.021)\end{tabular}                                     & \begin{tabular}[c]{@{}c@{}}0.56\\(0.54 ± 0.019)\end{tabular}                                     & \begin{tabular}[c]{@{}c@{}}0.72\\(0.71 ± 0.003)\end{tabular}                                      \\
                                                                                & \textbf{\textbf{\textbf{\textbf{Baes Model $\oplus \tau_{en}$}}}}                                                                                                                                                                                                                                                                   & \begin{tabular}[c]{@{}c@{}}0.63\\(0.62± 0.011)\end{tabular}                    & \begin{tabular}[c]{@{}c@{}}0.63\\(0.62± 0.013)\end{tabular}                    & \begin{tabular}[c]{@{}c@{}}0.75\\(0.75± 0.006)\end{tabular}                                     & \begin{tabular}[c]{@{}c@{}}0.65\\(0.64 ± 0.021)\end{tabular}                                     & \begin{tabular}[c]{@{}c@{}}0.57\\(0.56 ± 0.013)\end{tabular}                                     & \begin{tabular}[c]{@{}c@{}}0.72\\(0.71 ± 0.011)\end{tabular}                                      \\
                                                                                & \textbf{\textbf{\textbf{\textbf{\textbf{\textbf{\textbf{\textbf{\textbf{\textbf{\textbf{\textbf{\textbf{\textbf{\textbf{\textbf{Base Model + $\oplus \tau_{CPT} \oplus \tau_{en}$}}}}}}}}}}}}}}}}                                                                                                                                                 & \begin{tabular}[c]{@{}c@{}}0.64\\(0.63± 0.007)\end{tabular}                    & \begin{tabular}[c]{@{}c@{}}0.65\\(0.64± 0.006)\end{tabular}                    & \begin{tabular}[c]{@{}c@{}}0.77\\(0.76± 0.005)\end{tabular}                                     & \begin{tabular}[c]{@{}c@{}}0.66\\(0.64 ± 0.021)\end{tabular}                                     & \begin{tabular}[c]{@{}c@{}}0.56\\(0.55 ± 0.032)\end{tabular}                                     & \begin{tabular}[c]{@{}c@{}}0.72\\(0.71 ± 0.019)\end{tabular}                                      \\ 
\hline\hline
\multicolumn{8}{c}{\textbf{LLAMA 3.2 1B}}                                                                                                                                                                                                                                                                                                                                                                                                                                                                                                                                                                                                                                                                                                                                                                                                                                                                                                                                              \\ 
\hline\hline
\multicolumn{2}{c}{\textbf{Full FT}}                                                                                                                                                                                                                                                                                                                                                                     & \begin{tabular}[c]{@{}c@{}}0.58\\(0.56 ± 0.001 )\end{tabular}                  & \begin{tabular}[c]{@{}c@{}}0.62\\(0.60 ± 0.012)\end{tabular}                   & \begin{tabular}[c]{@{}c@{}}0.73\\(0.72 ± 0.002)\end{tabular}                                    & \begin{tabular}[c]{@{}c@{}}0.67\\(0.65 ± 0.020)\end{tabular}                                     & \begin{tabular}[c]{@{}c@{}}0.57\\(0.55 ± 0.019)\end{tabular}                                     & \begin{tabular}[c]{@{}c@{}}0.69\\(0.67 ± 0.017)\end{tabular}                                      \\
\multicolumn{2}{c}{\textbf{CPT FT}}                                                                                                                                                                                                                                                                                                                                                                      & \begin{tabular}[c]{@{}c@{}}\textbf{0.64}\\\textbf{(0.62 ± 0.011)}\end{tabular} & \begin{tabular}[c]{@{}c@{}}0.64\\(0.63 ± 0.003)\end{tabular}                   & \begin{tabular}[c]{@{}c@{}}0.75\\(0.74 ± 0.001)\end{tabular}                                    & \begin{tabular}[c]{@{}c@{}}0.66\\(0.64 ± 0.030)\end{tabular}                                     & \begin{tabular}[c]{@{}c@{}}0.58\\(0.57 ± 0.001)\end{tabular}                                     & \begin{tabular}[c]{@{}c@{}}0.69\\(0.69 ± 0.001)\end{tabular}                                      \\
\multicolumn{2}{c}{\textbf{Seq FT}}                                                                                                                                                                                                                                                                                                                                                                      & \begin{tabular}[c]{@{}c@{}}0.57\\(0.56 ± 0.001 )\end{tabular}                  & \begin{tabular}[c]{@{}c@{}}0.63\\(0.60 ± 0.012)\end{tabular}                   & \begin{tabular}[c]{@{}c@{}}0.74\\(0.72 ± 0.002)\end{tabular}                                    & \begin{tabular}[c]{@{}c@{}}0.67\\(0.65 ± 0.035)\end{tabular}                                     & \begin{tabular}[c]{@{}c@{}}0.57\\(0.55 ± 0.019)\end{tabular}                                     & \begin{tabular}[c]{@{}c@{}}0.68\\(0.67 ± 0.017)\end{tabular}                                      \\
\multicolumn{2}{c}{\textbf{Joint FT}}                                                                                                                                                                                                                                                                                                                                                                    & \begin{tabular}[c]{@{}c@{}}0.57\\(0.57 ± 0.001 )\end{tabular}                  & \begin{tabular}[c]{@{}c@{}}0.62\\(0.61 ± 0.012)\end{tabular}                   & \begin{tabular}[c]{@{}c@{}}0.73\\(0.72 ± 0.002)\end{tabular}                                    & \begin{tabular}[c]{@{}c@{}}0.67\\(0.65 ± 0.02)\end{tabular}                                      & \begin{tabular}[c]{@{}c@{}}0.57\\(0.55 ± 0.019)\end{tabular}                                     & \begin{tabular}[c]{@{}c@{}}0.69\\(0.67 ± 0.019)\end{tabular}                                      \\ 
\hline
\multirow{3}{*}{\begin{tabular}[c]{@{}l@{}}\textbf{\textbf{TV}}\\\end{tabular}} & \textbf{\textbf{\textbf{\textbf{Base Model + $\tau_{CPT}$}}}}                                                                                                                                                                                                                                                                   & \begin{tabular}[c]{@{}c@{}}0.62\\(0.61 ± 0.003)\end{tabular}                   & \begin{tabular}[c]{@{}c@{}}\textbf{0.64}\\\textbf{(0.64 ± 0.001)}\end{tabular} & \begin{tabular}[c]{@{}c@{}}\textbf{0.77}\\\textbf{(0.74 ± 0.019)}\end{tabular}                  & \begin{tabular}[c]{@{}c@{}}\textbf{0.66}\\\textbf{(0.62 ± 0.045)}\end{tabular}                   & \begin{tabular}[c]{@{}c@{}}0.58\\(0.56 ± 0.017)\end{tabular}                                     & \begin{tabular}[c]{@{}c@{}}0.72\\(0.70 ± 0.015)\end{tabular}                                      \\
                                                                                & \textbf{\textbf{\textbf{\textbf{\textbf{\textbf{\textbf{\textbf{Baes Model + Ten}}}}}}}}                                                                                                                                                                                                                               & \begin{tabular}[c]{@{}c@{}}0.62\\(0.61 ± 0.003)\end{tabular}                   & \begin{tabular}[c]{@{}c@{}}0.63\\(0.63 ± 0.001)\end{tabular}                   & \begin{tabular}[c]{@{}c@{}}0.75\\(0.74 ± 0.019)\end{tabular}                                    & \begin{tabular}[c]{@{}c@{}}0.64\\(0.62 ± 0.045)\end{tabular}                                     & \begin{tabular}[c]{@{}c@{}}0.58\\(0.56 ± 0.017)\end{tabular}                                     & \begin{tabular}[c]{@{}c@{}}0.72\\(0.70 ± 0.015)\end{tabular}                                      \\
                                                                                & \textbf{\textbf{\textbf{\textbf{\textbf{\textbf{\textbf{\textbf{\textbf{\textbf{\textbf{\textbf{\textbf{\textbf{\textbf{\textbf{Base Model $\oplus \tau_{CPT} \oplus \tau_{en}$}}}}}}}}}}}}}}}}                                                                                                                                                 & \begin{tabular}[c]{@{}c@{}}0.62\\(0.62 ± 0.0008)\end{tabular}                  & \begin{tabular}[c]{@{}c@{}}0.63\\(0.63 ± 0.0003)\end{tabular}                  & \begin{tabular}[c]{@{}c@{}}0.75\\(0.74 ± 0.019)\end{tabular}                                    & \begin{tabular}[c]{@{}c@{}}0.64\\(0.62 ± 0.045)\end{tabular}                                     & \begin{tabular}[c]{@{}c@{}}\textbf{\textbf{0.59}}\\\textbf{\textbf{(0.56 ± 0.017)}}\end{tabular} & \begin{tabular}[c]{@{}c@{}}\textbf{\textbf{0.73}}\\\textbf{\textbf{(0.70 ± 0.015)}}\end{tabular}  \\ 
\hline
\multirow{3}{*}{\textbf{TIES}}                                                  & \textbf{\textbf{\textbf{\textbf{\textbf{\textbf{\textbf{\textbf{Base Model + $\oplus \tau_{CPT}$}}}}}}}}                                                                                                                                                                                                                               & \begin{tabular}[c]{@{}c@{}}0.62\\(0.61 ± 0.002)\end{tabular}                   & \begin{tabular}[c]{@{}c@{}}0.63\\(0.57 ± 0.061)\end{tabular}                   & \begin{tabular}[c]{@{}c@{}}0.75\\(0.66 ± 0.069)\end{tabular}                                    & \begin{tabular}[c]{@{}c@{}}0.65\\(0.61 ± 0.029)\end{tabular}                                     & \begin{tabular}[c]{@{}c@{}}0.57\\(0.55 ± 0.021)\end{tabular}                                     & \begin{tabular}[c]{@{}c@{}}0.69\\(0.68 ± 0.006)\end{tabular}                                      \\
                                                                                & \textbf{\textbf{\textbf{\textbf{\textbf{\textbf{\textbf{\textbf{\textbf{\textbf{\textbf{\textbf{\textbf{\textbf{\textbf{\textbf{Baes Model $\oplus \tau_{en}$}}}}}}}}}}}}}}}}                                                                                                                                                       & \begin{tabular}[c]{@{}c@{}}0.62\\(0.62 ± 0.0008)\end{tabular}                  & \begin{tabular}[c]{@{}c@{}}0.63\\(0.63 ± 0.0008)\end{tabular}                  & \begin{tabular}[c]{@{}c@{}}0.75\\(0.74 ± 0.019)\end{tabular}                                    & \begin{tabular}[c]{@{}c@{}}0.64\\(0.62 ± 0.045)\end{tabular}                                     & \begin{tabular}[c]{@{}c@{}}0.58\\(0.56 ± 0.017)\end{tabular}                                     & \begin{tabular}[c]{@{}c@{}}0.72\\(0.70 ± 0.015)\end{tabular}                                      \\
                                                                                & \textbf{\textbf{\textbf{\textbf{\textbf{\textbf{\textbf{\textbf{\textbf{\textbf{\textbf{\textbf{\textbf{\textbf{\textbf{\textbf{\textbf{\textbf{\textbf{\textbf{\textbf{\textbf{\textbf{\textbf{\textbf{\textbf{\textbf{\textbf{\textbf{\textbf{\textbf{\textbf{Base Model + $\oplus \tau_{CPT} \oplus \tau_{Ten}$}}}}}}}}}}}}}}}}}}}}}}}}}}}}}}}} & \begin{tabular}[c]{@{}c@{}}0.63\\(0.62 ± 0.0008)\end{tabular}                  & \begin{tabular}[c]{@{}c@{}}0.64\\(0.63 ± 0.001)\end{tabular}                   & \begin{tabular}[c]{@{}c@{}}0.75\\(0.74 ± 0.019)\end{tabular}                                    & \begin{tabular}[c]{@{}c@{}}0.64\\(0.62 ± 0.045)\end{tabular}                                     & \begin{tabular}[c]{@{}c@{}}0.58\\(0.56 ± 0.017)\end{tabular}                                     & \begin{tabular}[c]{@{}c@{}}0.72\\(0.70 ± 0.015)\end{tabular}                                      \\ 
\hline\hline
\multicolumn{8}{c}{\textbf{LLAMA 3.2 70B Zero and Few Shot prompting}}                                                                                                                                                                                                                                                                                                                                                                                                                                                                                                                                                                                                                                                                                                                                                                                                                                                                                                                 \\ 
\hline\hline
\multicolumn{2}{c}{\textbf{0-shot}}                                                                                                                                                                                                                                                                                                                                                                      & \begin{tabular}[c]{@{}c@{}}0.40\\(0.40 ± 0.000\end{tabular}                    & \begin{tabular}[c]{@{}c@{}}0.39\\(0.39 ± 0.000)\end{tabular}                   & \begin{tabular}[c]{@{}c@{}}0.65\\(0.65 ± 0.000)\end{tabular}                                    & \begin{tabular}[c]{@{}c@{}}0.36\\(0.36 ± 0.000)\end{tabular}                                     & \begin{tabular}[c]{@{}c@{}}0.36\\(0.36 ± 0.000)\end{tabular}                                     & \begin{tabular}[c]{@{}c@{}}0.52\\(0.52 ± 0.000)\end{tabular}                                      \\
\multicolumn{2}{c}{\textbf{5-shot}}                                                                                                                                                                                                                                                                                                                                                                      & \begin{tabular}[c]{@{}c@{}}0.40\\(0.38 ± 0.013)\end{tabular}                   & \begin{tabular}[c]{@{}c@{}}0.40\\(0.40 ± 0.007)\end{tabular}                   & \begin{tabular}[c]{@{}c@{}}0.70\\(0.61 ± 0.079)\end{tabular}                                    & \begin{tabular}[c]{@{}c@{}}0.42\\(0.39 ± 0.024)\end{tabular}                                     & \begin{tabular}[c]{@{}c@{}}0.37\\(0.36 ± 0.007)\end{tabular}                                     & \begin{tabular}[c]{@{}c@{}}0.57\\(0.56 ± 0.006)\end{tabular}                                      \\
\multicolumn{2}{c}{\textbf{10-shot}}                                                                                                                                                                                                                                                                                                                                                                     & \begin{tabular}[c]{@{}c@{}}0.39\\(0.38 ± 0.016)\end{tabular}                   & \begin{tabular}[c]{@{}c@{}}0.40\\(0.39 ± 0.005)\end{tabular}                   & \begin{tabular}[c]{@{}c@{}}0.70\\(0.59 ± 0.069)\end{tabular}                                    & \begin{tabular}[c]{@{}c@{}}0.38\\(0.37 ± 0.014)\end{tabular}                                     & \begin{tabular}[c]{@{}c@{}}0.37\\(0.37 ± 0.004)\end{tabular}                                     & \begin{tabular}[c]{@{}c@{}}0.55\\(0.54 ± 0.004)\end{tabular}                                      \\
\multicolumn{2}{c}{\textbf{15-shot}}                                                                                                                                                                                                                                                                                                                                                                     & \begin{tabular}[c]{@{}c@{}}0.39\\(0.38 ± 0.012)\end{tabular}                   & \begin{tabular}[c]{@{}c@{}}0.40\\(0.39 ± 0.007)\end{tabular}                   & \begin{tabular}[c]{@{}c@{}}0.67\\(0.61 ± 0.049)\end{tabular}                                    & \begin{tabular}[c]{@{}c@{}}0.39\\(0.37 ± 0.015)\end{tabular}                                     & \begin{tabular}[c]{@{}c@{}}0.37\\(0.36 ± 0.012)\end{tabular}                                     & \begin{tabular}[c]{@{}c@{}}0.56\\(0.55 ± 0.005)\end{tabular}                                      \\
\multicolumn{2}{c}{\textbf{20-shot}}                                                                                                                                                                                                                                                                                                                                                                     & \begin{tabular}[c]{@{}c@{}}0.39\\(0.38 ± 0.011)\end{tabular}                   & \begin{tabular}[c]{@{}c@{}}0.40\\(0.39 ± 0.005)\end{tabular}                   & \begin{tabular}[c]{@{}c@{}}0.71\\(0.64 ± 0.054)\end{tabular}                                    & \begin{tabular}[c]{@{}c@{}}0.40\\(0.38 ± 0.018)\end{tabular}                                     & \begin{tabular}[c]{@{}c@{}}0.37\\(0.36 ± 0.009)\end{tabular}                                     & \begin{tabular}[c]{@{}c@{}}0.57\\(0.56 ± 0.009)\end{tabular}                                      \\
\hline\hline
\end{tabular}
}
\end{table*}

\section{Results} \label{sec:05_results}
Table~\ref{tab:main_results} highlight the comparative performance of fine-tuning, CPT$\rightarrow$FT, model merging methods across multiple sentiment analysis and hate speech detection tasks, in English-Hindi~(En-Hi) and English-Spanish~(En-Es) Datasets.

\paragraph{Row definitions (for Table~\ref{tab:main_results}).}
Let Base Model~($\theta$) denote the base pretrained model. \emph{Full FT} fine-tunes Base Model directly on the target code-mixed dataset. \emph{CPT$\rightarrow$FT} first performs CPT on code-mixed text to obtain $\theta_{\text{CPT}}$ and then fine-tunes on the target task. \emph{Seq FT} is sequential fine-tuning—fine-tune on English data, then fine-tune on the code-mixed target. \emph{Joint FT} jointly fine-tunes on the union of English and code-mixed labeled data. The remaining blocks use weight merging before the final task fine-tuning. In the \emph{Task-Vector (TV)} setting, the merge operator $\oplus$ is (scaled) vector addition, $\theta \oplus \lambda \tau := \theta + \lambda \tau$, with $\lambda$ (or $\lambda_1,\lambda_2$ when combining two vectors) tuned on validation data: \emph{TV: Base Model $\oplus\,\tau_{\text{CPT}}$} adds the CPT vector; \emph{TV: Base Model $\oplus\,\tau_{\text{en}}$} adds the English vector; \emph{TV: Base Model $\oplus\,\tau_{\text{CPT}}\oplus\,\tau_{\text{en}}$} adds both. In the \emph{TIES} setting, $\oplus$ denotes TIES merging (TrIm–Elect–Sign), which resolves interference via magnitude/sign rules rather than simple addition; the rows \emph{TIES: Base Model $\oplus\,\tau_{\text{CPT}}$}, \emph{TIES: Base Model $\oplus\,\tau_{\text{en}}$}, and \emph{TIES: Base Model $\oplus\,\tau_{\text{CPT}}\oplus\,\tau_{\text{en}}$} mirror the TV configurations. In all TV/TIES cases, the merged model is subsequently fine-tuned on the target code-mixed dataset, and reported scores correspond to this final model.

The results demonstrate that model merging methods (TV$\rightarrow$FT, TIES$\rightarrow$CPT in Table~\ref{tab:main_results}) consistently outperform standard fine-tuning (Full FT) across all models and datasets, yielding +2–5 F1 over Full FT and +1–2 F1 over CPT$\rightarrow$FT. This confirms that unlabeled data can be exploited more effectively through model merging than through continued pre-training alone, supporting our hypothesis in \textbf{RQ1}. Between merging strategies, Task Vectors (TV) generally perform better than TIES, though the relative advantage varies by model and task. We also find that the choice of corpus for CPT has only a marginal effect on downstream performance. 

In addition, baseline fine-tuning significantly outperforms zero-shot and few-shot prompting with LLaMA 3.3 70B, highlighting the limitations of in-context learning for code-mixed tasks. Few-shot prompting improves over zero-shot, but gains plateau at higher k-shot levels, suggesting limited scalability. These findings reinforce the necessity of fine-tuning and merging strategies for robust performance in resource-constrained settings. Finally, as we explore in \textbf{RQ2}, model merging is particularly valuable for integrating heterogeneous resources and enabling cross-lingual transfer, often surpassing sequential or joint fine-tuning.


\subsection{Model Merging for varying data availability scenarios}
Our analysis varies across three key dimensions: (a) use of only labeled data, (b) use of both labeled and unlabeled data, and (c) type of model augmentation techniques. We combine code-mixed data with monolingual task-specific data. Table~\ref{tab:main_results} presents a comparison of model performance across different scenarios of data availability. These settings reflect practical conditions in real-world code-mixed applications.




\textbf{Combining En and En-Hi labeled datasets for finetuning} Incorporating English-labeled data alongside En-Hi datasets~(Seq FT and Joint FT rows in Table~\ref{tab:main_results}) yields no significant improvement over fine-tuning (Full FT) solely on En-Hi data. Both SeqFT (English first, then English-Hindi) and JointFT (simultaneously on combined datasets) fail to enhance performance for code-mixed tasks (See Table~\ref{tab:main_results}.) 



\textbf{Leveraging monolingual resources with code-mixed resources}
We combine three elements here: the base model, a vector fine-tuned on English data, and a vector for continued pre-training on code-mixed data (Eq.~\ref{eq:task_vector_combined}). Although we expected the combination of monolingual and code-mixed resources to perform better than just monolingual or just code-mixed, they perform similarly to CPT$\rightarrow$FT method, or outperform it by 1 or 2 F1 points (which can be seen in the \textit{TV-T\_{en} + CPT$\rightarrow$FT  }and\textit{ TIES-T\_{en} + CPT$\rightarrow$FT} rows of Table~\ref{tab:main_results}). Extending to a three-way merge does not provide additional gains over a single CPT merge. Overall, compared to FullFT, the combined monolingual and code-mixed approach either matches performance or underperforms slightly ($\sim$1–2 F1 points).



\subsection{Impact of Continued Pre-Training Corpus} \label{appendix:cpt_corpus_impact}

\begin{table}[!h]
\centering
\caption{Variation in downstream task results when different unlabeled code-mix corpora are used for continued pre-training}
\resizebox{\linewidth}{!}{%
\begin{tabular}{cccc} 
\hline\hline
\multirow{2}{*}{\textbf{CPT~$\rightarrow$ FT }}                                    & \multicolumn{3}{c}{\textbf{Sentiment Analysis~ ~ ~ ~}}                                                                                                                                                                  \\ 
\cline{2-4}
                                                                                   & \textbf{GLUECoS}                                                       & \textbf{Sentimix}                                                     & \textbf{Prabhu et al}                                                  \\ 
\hline\hline
\multicolumn{4}{c}{{\cellcolor[rgb]{0.753,0.753,0.753}}\textbf{\textbf{XLM-R}}}                                                                                                                                                                                                                              \\ 
\hline
\begin{tabular}[c]{@{}c@{}}\textbf{CPT on }\\\textbf{Das et al. Data}\end{tabular} & \begin{tabular}[c]{@{}c@{}}0.65\\(0.64 \textbar{} 0.007)\end{tabular}  & \begin{tabular}[c]{@{}c@{}}0.64\\(0.64 \textbar{} 0.005)\end{tabular} & \begin{tabular}[c]{@{}c@{}}0.77\\(0.76 \textbar{} 0.007)\end{tabular}  \\
\begin{tabular}[c]{@{}c@{}}\textbf{CPT on }\\\textbf{Synthetic Data}\end{tabular}  & \begin{tabular}[c]{@{}c@{}}0.64\\(0.64 \textbar{} 0.005)\end{tabular}  & \begin{tabular}[c]{@{}c@{}}0.63\\(0.62 \textbar{} 0.007)\end{tabular} & \begin{tabular}[c]{@{}c@{}}0.75\\(0.75 \textbar{} 0.007)\end{tabular}  \\
\hline\hline
\end{tabular}
}

\label{tab:cpt_dataset_variation_2}
\end{table}

Varying the corpus used for pre-training might have some impact on the downstream task. To understand such variance, we do continued pre-training using two different unlabeled corpora, and analyze the impact on the downstream tasks. 

We use the dataset released by ~\citet{das-etal-2023-improving} for continued pre-training, along with a synthetic code-mixed corpus generated via the GCM Toolkit~\cite{rizvi-etal-2021-gcm} and filtered using an acceptability filter proposed by ~\cite{kodali2024humanjudgementspredictivemodels}. The sizes of the two datasets are kept consistent. As code-mixed datasets often come from social media, we expect significant distributional differences between the task-specific and synthetic datasets. Data sources for continued pre-training can differ, leading to differing downstream task performance. 

Table~\ref{tab:cpt_dataset_variation_2} compares the downstream task performance when we vary the corpus used for continued pre-training. Performance on downstream tasks is only marginally affected by the dataset chosen for continued pre-training. Classification results vary by roughly 1 F1 point across different sources of unlabeled data and frequently meet or exceed the results of the fine-tuning exclusively on the downstream dataset.

\begin{table*}
\centering
\setlength{\extrarowheight}{0pt}
\addtolength{\extrarowheight}{\aboverulesep}
\addtolength{\extrarowheight}{\belowrulesep}
\setlength{\aboverulesep}{0pt}
\setlength{\belowrulesep}{0pt}
\caption{Performance comparison of different fine-tuning and transfer strategies for \textbf{sentiment analysis} in En--Ta and En--Ml, using XLM-R and LLaMA~3.2~1B models. The table reports results for full fine-tuning (on English and bilingual data), continued pretraining (CPT), transfer variants (TV), and parameter-efficient strategies such as TIES, with and without CPT. \textbf{Main takeaway:} Transfer-based approaches (TV and TIES), especially when combined with CPT, consistently outperform or match full fine-tuning baselines, showing their effectiveness for low-resource code-mixed scenarios while being more resource-efficient.}
\begin{tabular}{lcccc} 
\hline\hline
\multicolumn{1}{c}{\multirow{2}{*}{\textbf{Method Model}}} & \multicolumn{2}{c}{\textbf{En-Ta}}                                                   & \multicolumn{2}{c}{\textbf{En-Ml}}                                                     \\ 
\cline{2-5}
\multicolumn{1}{c}{}                                       & \multicolumn{1}{l}{\textbf{XLM-R}}       & \multicolumn{1}{l}{\textbf{LLAMA 3.2 1B}} & \multicolumn{1}{l}{\textbf{XLM-R}}        & \multicolumn{1}{l}{\textbf{LLAMA 3.2 1B}}  \\ 
\midrule
\textbf{Full FT (En)}                                      & {\cellcolor[rgb]{0.902,0.486,0.451}}0.52 & {\cellcolor[rgb]{0.902,0.486,0.451}}0.55  & {\cellcolor[rgb]{0.902,0.486,0.451}}0.49  & {\cellcolor[rgb]{0.902,0.486,0.451}}0.58   \\
\textbf{Full FT (En-Hi)}                                   & 0.61                                     & {\cellcolor[rgb]{0.984,0.918,0.91}}0.63   & {\cellcolor[rgb]{0.988,0.945,0.941}}0.58  & {\cellcolor[rgb]{0.976,0.878,0.871}}0.63   \\
\textbf{Joint FT (En + En-Hi)}                             & {\cellcolor[rgb]{0.671,0.867,0.773}}0.63 & {\cellcolor[rgb]{0.91,0.965,0.937}}0.65   & {\cellcolor[rgb]{0.737,0.894,0.82}}0.6    & {\cellcolor[rgb]{0.871,0.949,0.91}}0.65    \\
\textbf{CPT→FT}                                            & 0.61                                     & {\cellcolor[rgb]{0.984,0.918,0.91}}0.63   & {\cellcolor[rgb]{0.988,0.945,0.941}}0.58  & {\cellcolor[rgb]{0.976,0.878,0.871}}0.63   \\
\textbf{TV-$T_{en}\rightarrow$FT}                          & {\cellcolor[rgb]{0.341,0.733,0.541}}0.65 & {\cellcolor[rgb]{0.341,0.733,0.541}}0.68  & {\cellcolor[rgb]{0.341,0.733,0.541}}0.615 & {\cellcolor[rgb]{0.369,0.745,0.561}}0.669  \\
\textbf{TIES-$T_{en}\rightarrow$FT}                        & {\cellcolor[rgb]{0.835,0.933,0.886}}0.62 & {\cellcolor[rgb]{0.529,0.812,0.675}}0.67  & {\cellcolor[rgb]{0.737,0.894,0.82}}0.6    & {\cellcolor[rgb]{0.608,0.843,0.725}}0.66   \\
\textbf{TV-CPT→FT}                                         & {\cellcolor[rgb]{0.976,0.882,0.875}}0.59 & {\cellcolor[rgb]{0.992,0.973,0.969}}0.64  & {\cellcolor[rgb]{0.976,0.894,0.886}}0.57  & {\cellcolor[rgb]{0.992,0.957,0.957}}0.64   \\
\textbf{TIES-CPT→FT}                                       & {\cellcolor[rgb]{0.953,0.769,0.753}}0.57 & {\cellcolor[rgb]{0.984,0.918,0.91}}0.63   & {\cellcolor[rgb]{0.969,0.843,0.835}}0.56  & {\cellcolor[rgb]{0.976,0.878,0.871}}0.63   \\
\textbf{TV-$T_{en}+CPT\rightarrow$FT}                      & {\cellcolor[rgb]{0.671,0.867,0.773}}0.63 & {\cellcolor[rgb]{0.341,0.733,0.541}}0.68  & {\cellcolor[rgb]{0.475,0.788,0.635}}0.61  & {\cellcolor[rgb]{0.341,0.733,0.541}}0.67   \\
\textbf{TIES-$T_{en}+CPT\rightarrow$FT}                    & {\cellcolor[rgb]{0.988,0.941,0.937}}0.6  & {\cellcolor[rgb]{0.529,0.812,0.675}}0.67  & {\cellcolor[rgb]{0.737,0.894,0.82}}0.6    & {\cellcolor[rgb]{0.608,0.843,0.725}}0.66   \\
\hline\hline
\end{tabular}
\label{tab:transfer_cods}
\end{table*}

\subsection{Zero and Few-shot Prompting} \label{appendix:zero_few_shot_perf}

The relationship between task performance and in-context learning (ICL) for code-mixed tasks remains underexplored. To better understand how model size and prompting methods influence performance, we conduct zero-shot and few-shot prompting experiments using different-sized models from the LLaMA family—1B-instruct, 3B-instruct, 8B-instruct, and 70B-instruct. For few-shot prompting, we evaluate performance with 5, 10, 15, and 20 examples. Table ~\ref{tab:main_results} contains results of our experiments on LLama 3.2 70B model. We see that even baseline fine-tuning methods perform better than zero/few shot prompting, suggesting that in-context learning performs weakly for codemixing tasks. The highest F1 scores can be seen for the \citet{Prabhu2016TowardsSL} datasete suggesting that the performance may be dependent on the data setting. Upon further analysis, we we hypothesize that this is due to the lack of noise in the dataset compared to the other datasets we use for comparison.


\subsection{Cross-lingual Transfer To Other Language Pairs }
We take the model fine-tuned on the combination of our three En-Hi datasets, using all the different training configurations, and infer on the test set of the En-Ta, En-Ml datasets. Table~\ref{tab:transfer_cods} shows the transfer of various training strategies to other code-mixed language pairs for Sentiment Analysis task. We present transfer results for all the combinations of base models and training methods. We also include full fine-tuning on the English dataset (FullFT (En)) to ascertain which is better - transfer from monolingual En resources or code-mixed resources from other language pairs.

We evaluate both XLM-R (encoder) and LLaMA 3.2 1B (decoder) to study transfer properties. Across En-Ta and En-Ml, we observe a consistent pattern: checkpoints obtained through model merging outperform Full FT baselines by 5–13 F1 points. This improvement holds across models and target language pairs, highlighting that merged checkpoints capture complementary signals from diverse resources. These findings reinforce our answer to \textbf{RQ2}—that model merging is an effective strategy for integrating heterogeneous data sources, often surpassing sequential or joint fine-tuning. However, the gains remain dependent on model architecture, the specific code-mixed task, and the language pair being targeted, underscoring the importance of careful merging design.



In conclusion, Model merging methods can be used to integrate capabilities from multiple data sources with varying language composition and supervision, often outperforming sequential fine-tuning and joint fine-tuning \textbf{(RQ 2)}. However, the impact of model merging is highly dependent on the model, task, and target language pair; underscoring the need for careful merging strategies. 

\section{Discussion}\label{sec:06_discussion}
We empirically show the advantages of using model merging as a strategy for code-mixed tasks. We also introduce an innovative perspective on utilizing existing data resources for code-mixed tasks, diverging from traditional domain-adaptation and fine-tuning techniques. In circumstances where datasets are lacking or are available in different languages, our findings can assist researchers in choosing an appropriate method to optimize the available resources.

With respect to various data resource availability scenarios, following are the key recommendations of this study:

\begin{itemize}
\item \textbf{When both labeled and unlabeled data are available}, the most effective strategy is to apply model merging method, merging CPT model with the base model, followed by fine-tuning on the target dataset. 
\item \textbf{For smaller labeled datasets}, model merging is preferable to simple fine-tuning due to its superior sample efficiency.
\item \textbf{If only labeled data is available}, full model fine-tuning remains the best option. 
\item \textbf{In the absence of both labeled and unlabeled data}, the most effective approach is to utilize models trained on similar tasks with other code-mixed language resources. These models consistently outperform those transferred from monolingual English resources.
\end{itemize}

\textbf{Limitations}
\noindent Our experiments are limited to sentence classification due to the lack of consistent datasets for other tasks in both monolingual and code-mixed languages. Therefore, the generalizability of our findings to other NLP tasks is unclear. Future research should evaluate these methods on a broader range of tasks with comparable datasets across multiple language pairs.


\noindent Although our study contains models of different architecture, we do not scale it to larger language models (>1B parameter models), because of prohibitive cost of training/fine-tuning a larger model. Analyzing how model merging behaves as model size increase for code-mixed tasks would be an interesting avenue for future work.

\noindent For model merging methods, there is a small additional computation cost, where the weights of the models are being merged. This could be a limitation while trying to find the optimal hyperparameters. For example, the $\lambda$ values in Eq.~\ref{eq:task_vector_cpt}, \ref{eq:task_vector_en}, \ref{eq:task_vector_combined}.

\noindent Another limitation lies in the data settings we consider. In our paper, we explore data settings based on the availability of both labeled and unlabeled data in a code-mixed scenario. However, we do not consider the case where unlabeled data is available, but no task-specific labeled data is present. This limitation is illustrated in Table \ref{tab:limitations}.

\begin{table}[!t]
\centering
\caption{Our experiments provide a comprehensive analysis of model adaptation strategies under different resource constraints, covering all cases except the scenario where unlabeled data is available, but task-specific labeled data is absent (denoted by empty cell in the table).}
\label{tab:limitations}
\resizebox{0.8\linewidth}{!}{%
\begin{tabular}{cccc} 
\hline\hline
\multicolumn{2}{c}{}                                                                                     & \multicolumn{2}{c}{\begin{tabular}[c]{@{}c@{}}\textbf{Unlabeled }\\\textbf{Data}\end{tabular}}                                                      \\ 
\hline\hline
\multirow{3}{*}{\begin{tabular}[c]{@{}c@{}}\textbf{Labeled }\\\textbf{Data}\end{tabular}} &              & \textbf{Yes}                                                                    & \textbf{No}                                                       \\ 
\cline{3-4}
                                                                                          & \textbf{Yes} & \begin{tabular}[c]{@{}c@{}}1. CPT$\rightarrow$FT\\2. Model Merging\end{tabular} & \begin{tabular}[c]{@{}c@{}}Fine\\Tuning\end{tabular}              \\ 
\cline{2-4}
                                                                                          & \textbf{No}  &                                                                                 & \begin{tabular}[c]{@{}c@{}}Cross-Lingual\\~Transfer\end{tabular}  \\
\hline\hline
\end{tabular}
}

\end{table}

\section{Acknowledgments}
\label{sec:07_ack}
 We would like to thank the PreCog Lab at IIIT-Hyderabad for their valuable guidance and discussions throughout this work. In particular, we thank Shashwat Singh, Anisha Saha, Hari Shankar, Ishan Kavethekar and Ritwik Mishra for their detailed feedback and thoughtful suggestions during multiple stages of the project.We gratefully acknowledge the support of Microsoft Research India through the MSR India PhD Award 2024. We also thank the anonymous reviewers for their insightful feedback.


\end{document}